\RequirePackage{fix-cm}

\documentclass[twocolumn]{svjour3}   

\smartqed  

\usepackage{graphicx,color,multirow,booktabs,bigstrut}

\usepackage[misc]{ifsym}

\usepackage{xpatch,color}

\usepackage[noend]{algpseudocode}

\usepackage{algorithmicx,algorithm}

\usepackage{float}

\begin{document}

\title{Abs-CAM: A Gradient Optimization Interpretable Approach for Explanation of Convolutional Neural Networks}

\author{Chunyan Zeng\textsuperscript{1} \and
		Kang Yan\textsuperscript{1} \and
        Zhifeng Wang\textsuperscript{2} \and
        Yan Yu\textsuperscript{1} \and
        Shiyan Xia\textsuperscript{1} \and
        Nan Zhao\textsuperscript{1}
        }

\institute{\Letter Zhifeng Wang  \at
               \email{zfwang@ccnu.edu.cn}           \\       
   \at
    {$^1$} Hubei Key Laboratory for High-efficiency Utilization of Solar Energy and Operation Control of Energy Storage System, Hubei University of Technology, Wuhan 430068, China 
    \at   
    {$^2$} Department of Digital Media Technology, Central China Normal University, Wuhan 430079, China
}

\date{Received: 8 July 2022}

\maketitle

\begin{abstract}
The black-box nature of Deep Neural Networks (DNNs) severely hinders its performance improvement and application in specific scenes. In recent years, class activation mapping-based method has been widely used to interpret the internal decisions of models in computer vision tasks. However, when this method uses backpropagation to obtain gradients, it will cause noise in the saliency map, and even locate features that are irrelevant to decisions. In this paper, we propose an Absolute value Class Activation Mapping-based (Abs-CAM) method, which optimizes the gradients derived from the backpropagation and turns all of them into positive gradients to enhance the visual features of output neurons' activation, and improve the localization ability of the saliency map. The framework of Abs-CAM is divided into two phases: generating initial saliency map and generating final saliency map. The first phase improves the localization ability of the saliency map by optimizing the gradient, and the second phase linearly combines the initial saliency map with the original image to enhance the semantic information of the saliency map. We conduct qualitative and quantitative evaluation of the proposed method, including Deletion, Insertion, and Pointing Game. The experimental results show that the Abs-CAM can obviously eliminate the noise in the saliency map, and can better locate the features related to decisions, and is superior to the previous methods in recognition and localization tasks.
\end{abstract}

\begin{table*}[htbp]
	\centering
	\caption{Summary of different interpretable methods.}
	\begin{tabular}{r|l|l}
		\hline
		\multicolumn{1}{l|}{\textbf{Category}} & \textbf{Methods} & \textbf{Characteristics} \bigstrut\\
		\hline
		& RISE [5] & Simple, weak interpretation effect \bigstrut\\
		\cline{2-3}    \multicolumn{1}{l|}{Perturbation-based} & Fong et al. [6] & Effectively block important pixels \bigstrut\\
		\cline{2-3}          & Agarwal et al. [7] & Natural perturbed image, time-consuming \bigstrut\\
		\hline
		& Deconvolution [8] & Display target contour, noisy \bigstrut\\
		\cline{2-3}    \multicolumn{1}{l|}{Backpropagation-based} & Integrated Gradient [9] & Good performance on neuron saturation \bigstrut\\
		\cline{2-3}          & Smilkov et al. [10] & Requires multiple iterations, time-consuming \bigstrut\\
		\hline
		& CAM [11] & Retrain the model, time-consuming \bigstrut\\
		\cline{2-3}          & Grad-CAM [1] & Class activation maps are noisy \bigstrut\\
		\cline{2-3}          & Grad-CAM++ [2] & Good visualization effect \bigstrut\\
		\cline{2-3}    \multicolumn{1}{l|}{Class activation mapping-based} & Score-CAM [3] & Less noise in class activation map \bigstrut\\
		\cline{2-3}          & Group-CAM [12] & Generate the saliency map efficiently \bigstrut\\
		\cline{2-3}          & LFI-CAM [4] & Improve classification performance \bigstrut\\
		\cline{2-3}          & Relevance-CAM [13] & Works effectively at any layer \bigstrut\\
		\hline
	\end{tabular}%
	\label{tab:addlabel}%
\end{table*}%

\section{Introduction}
DNNs have shown superior performance on computer vision tasks. A key factor for the success of DNNs is that its network is deep enough, and the complex combination of a large number of nonlinear network layers can extract features at various levels of abstraction from the original data. However, the characteristics of DNNs such as many parameters, "end-to-end", and complex combinations of a large number of nonlinear network layers make it impossible to understand the mechanism of the model's decision-making. The lack of model interpretability severely hinders its application in high-risk decision-making fields such as medical diagnosis, finance, autonomous driving, and military. Therefore, the research on the interpretability of DNNs is of great significance.

In order to improve model transparency and make people trust the model, many interpretability methods have been proposed. Among them, the visualization method is widely studied, it interprets the model representation and decision-making in the form of a saliency map, where the intensity of the pixel colors in the saliency map corresponds to the importance of the decision result. Class activation mapping-based interpretable method is the most advanced visualization saliency map technology. According to the different ways of weight acquisition, this category of method is further divided into two categories: gradient-based methods and gradient-free methods. Gradient-based Class Activation Mapping (CAM) methods include Gradient-weighted Class Activation Mapping (Grad-CAM) \cite{selvaraju2017grad}, Grad-CAM++ \cite{chattopadhay2018grad}, etc. Gradient-free Class Activation Mapping methods include Score-CAM \cite{wang2020score}, Learning Feature Importance Class Activation Mapping (LFI-CAM) \cite{2021LFI}, etc. Gradient-based class activation mapping methods obtain weights by backpropagation, and then linearly combine with the activation map to generate a saliency map to visualize the region of interest of Convolutional Neural Networks (CNNs). The saliency maps have better class-discriminative, but also have the shortcomings of roughness, noisy, and will locate to regions irrelevant to the decision.

In order to improve the localization ability and enhance the semantic information of the saliency map, we propose a new CAM method, Abs-CAM. The method is inspired by Grad-CAM \cite{selvaraju2017grad} and optimizes the gradients obtained by backpropagation derivation, and adds the absolute value to turn all the gradients into positive gradients, which enhance its feature localization ability. And inspired by the frameworks of RISE \cite{petsiuk2018rise}  and Score-CAM \cite{wang2020score}, the framework of Abs-CAM includes two phases: generating initial saliency map and generating final saliency map. The first phase improves the localization ability of the saliency map by optimizing the gradient, and the second phase linearly combines the initial saliency map with the original image to enhance the semantic information of the saliency map. Our contributions are summarized as follows:

(1). We propose a novel gradient optimization interpretable method, Abs-CAM, whose framework is divided into two phases: generating initial saliency map and generating final sali\-ency map, the generated initial saliency map is point multiplied with the original image and then fed into the model. It further optimizes the gradients so that the target region of interest is located more intensively, with only a small amount of redundant information at the edges of the region.

(2). The visual effect of Abs-CAM is qualitatively evaluated through experiments. The saliency map generated by this method is smoother, the noise is obviously removed, and the features related to decision-making can be better located.

(3). Abs-CAM is quantitatively evaluated by evaluation indicators such as Deletion, Insertion and Pointing Game. Compared with other class activation mappi\-ng-based methods, our method is superior in explaining CNNs’ decisions.

The rest of the paper is organized as follows: Section \ref{2} gives an overview of the related work in the area of visualization interpretable methods. Section \ref{3} describes the proposed method by a step-by-step detailed explanation. Experiments are presented in Section \ref{4}, while the conclusions are given in Section \ref{5}.

\section{Related work} \label{2}
According to the time when interpretability is obtained, interpretability methods are divided into intrinsic interpretation and post-hoc interpretation. Further, according to the different scope of interpretation, post-hoc interpretation can be divided into global interpretation and local interpretation. We can understand how the model makes predictions as a whole through global interpretation, while local interpretation provides local-scope explanations for the model's individual predictions. Perturbation-based, backpropagation-based, and class activation mapping-based local interpretation methods have been widely studied among those interpretability methods. The analysis of these three types of methods in this paper is shown in Table 1.

\subsection{Perturbation-based methods}
Perturbation-based methods examine the causal relationship between input and output, it determines the impact of input disturbances on output by repeatedly perturbing the input and then observing the changes mapped to the output. The perturbation methods are classified as occluding, blurring, adding noise, and generative perturbation, etc. In RISE \cite{petsiuk2018rise}, Petsiuk et al. use Monte Carlo sampling to generate masks to occlude the input, and then compute the weighted sum of the masks and prediction scores to obtain the final saliency map. RISE adopts a simple perturbation method, and the interpretation process is simple, but the interpretation effect is poor. Subsequently, Fong et al. \cite{fong2017interpretable} argue that the perturbation masks can be learned, and the perturbation masks learned by optimization ideas can occlude important pixels more effectively than masks generated by simple perturbation. Agarwal et al. \cite{agarwal2019removing} use generative models to perturb the input, and it can obtain more natural perturbed images in the visual. However, this perturbation method requires multiple optimization iterations and takes a long time. The saliency map generated by the perturbation-based method is smoother, but there is more redundant information on the edge of the target region.

\subsection{Backpropagation-based methods}
Backpropagation-based methods use backpropagation to compute the gradient of the output of the model relative to the input, thereby highlighting the image regions that have a greater impact on the prediction. Zeiler et al. \cite{zeiler2014visualizing} use deconvolution to explain what each layer of the CNN learns. The visualization shows the target outline, but there is a lot of noise. Subsequently, Sundararajan et al. \cite{sundararajan2016gradients} propose the Integrated Gradient, which quantifies the importance of each component of the input by integrating the gradient. It effectively solves the problem of neuron saturation in DNNs, which leads to the inability to use gradient information to reflect the importance of features. Smilkov et al. \cite{smilkov2017smoothgrad} add Gaussian noise to the input image to achieve the smoothing and denoising gradient maps, but this method requires multiple iterations and takes a long time. Backpropagation-based methods can effectively locate the decision features of the input image, but there is clearly visible noise in the saliency map, while the gradient information can only be used to locate important features and cannot quantify the degree of contribution of each feature to the decisions.

\subsection{Class activation mapping-based methods}
Class activation mapping-based methods visualize the area that CNN focuses on by linearly combines weights and activation maps to generate saliency maps. According to the different ways of weight acquisition, these methods are further divided into two categories: gradient-based and gradient-free. Initially, Zhou et al. \cite{zhou2016learning} propose the CAM, in which the weights of the output layer are expressed as the score values of the predicted classes. CAM requires modifying the model network structure and retraining the model, which will consume a lot of time. Subsequently, in order to solve this problem, Grad-CAM \cite{selvaraju2017grad} is proposed. It calculates the gradient of the target class score relative to each feature map in the last convolutional layer and obtains the weights by global averaging of the gradients. Grad-CAM works with various CNNs models without modifying the model structure or retraining the model. But the class activation maps generated by Grad-CAM are significantly noisy and contain more redundant information at the edges of the regions, providing only coarse-grained interpretation of the results. Grad-CAM++ \cite{chattopadhay2018grad} is the same as Grad-CAM in that the weights are calculated by backpropagation. The difference is that the Grad-CAM++ calculates the weighted combination of the positive partial derivatives of the target class score relative to the last convolutional layer feature maps as weights, which effectively improves the visualization of the image. Unlike the above methods, Score-CAM \cite{wang2020score} uses the gradient-free approach to obtain the weights of each channel, it can generate a class activation map with less noise and better visual effect. In order to reduce calculation time of the saliency map, Group-CAM \cite{2021Group} uses the "split-transform-merge" strategy to generate the saliency maps efficiently and effectively, but its saliency maps have some noise. LFI-CAM \cite{2021LFI} combines Attention Branch Network (ABN) to improve the accuracy and classification performance of CAM. Relevance-CAM \cite{Lee_2021_CVPR} utilizes layer-wise relevance propagation to obtain the weighting components, it works effectively at any layer. The class activation map generated by class activation mapping-based methods has good class discrimination, but it also has disadvantages such as roughness, noise, and locating features that are not related to decision-making.

\begin{figure*}[h!]
	\centering
	\includegraphics[width=5.8in]{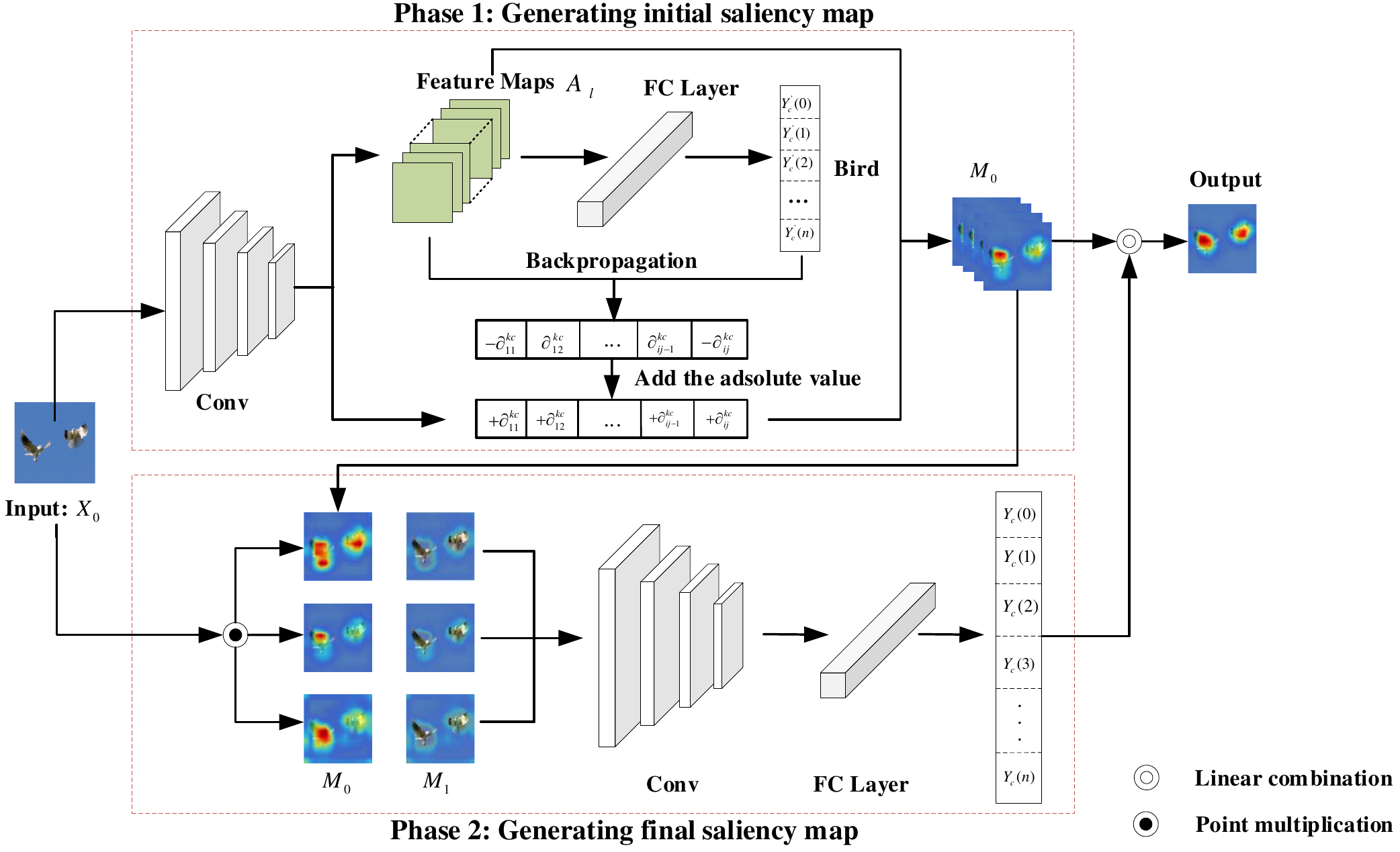}
	\caption{The framework of proposed Absolute value Class Activation Mapping-based interpretable approach.}
	\label{fig1}
\end{figure*}

\section{Methodology} \label{3}
The framework of proposed Abs-CAM is shown in Fig. \ref{fig1}. In the first phase, the gradients are optimized and all of them are changed to positive values to obtain the initial saliency map. In the second phase, the initial saliency map is multiplied with the input image and fed into the model to obtain the final saliency map.

\subsection{Initial saliency map} 
The generic model ${Y}$ analyzed in this paper is a neural network for predictive classification. For a given input image ${X_0}$, model ${Y}$ can outputs a score $Y_{c}^{\prime}\left(X_{0}\right)$ that predicts the class $c$ of that image. ${A_l}$  is denoted as the feature maps of the target convolutional layer ${l}$ output in ${Y}$. The ${k}$-th feature map of
${A_l}$ is denoted as ${A^k_l}$. We first compute the gradient $\partial _{ij}^{kc}$ of the score $Y_{c}^{\prime}\left(X_{0}\right)$ with respect to the spatial location (${i}$, ${j}$) in the feature map ${A_l^k}$.

\begin{equation}
	\centering
{\partial _{ij}^{kc}} = \frac{{\partial Y_{c}^{\prime}\left(X_{0}\right) }}{{\partial A_l^k}}
\end{equation}

Then optimize the gradient $\partial _{ij}^{kc}$, adds the absolute value to turn all the gradients into positive gradients and obtain the gradient $|\partial _{ij}^{kc}|$.

%\begin{equation}
%{\partial _k^c} = |\partial _{ij}^{kc}|
%\end{equation}

These gradients flowing back are global-average-poo\-led to obtain the weight $W_k^c$.

\begin{equation}
{W_k^c} = \frac{1}{Z}\sum\limits_{i} {\sum\limits_{j} {|\partial _{ij}^{kc}|} }
\end{equation}
where ${Z}$ represents the number of pixels of the feature map  ${A_l^k}$.

Finally, the weights are multiplied with the corresponding feature maps, and the up-sampling and normalization operations are performed to obtain the initial saliency map ${M_0}$.

\begin{equation}
{M_0} = S\left( {U\left( {A_l^k\cdot W_k^c} \right)} \right)
\end{equation}
where ${U}$ represents up-sampling operation and ${S}$ represents normalization operation.

\subsection{Final Saliency Map}
The second phase of the processing is performed in order to enhance the visual effect and semantic information of the saliency map. First, the initial saliency map ${M_0}$ is multiplied pointwise with the input image ${X_0}$ to obtain ${M_1}$.

\begin{equation}
{M_1} = {M_0}\cdot{X_0}
\end{equation}

Then it is fed into the model and after softmax to get ${Y_c}\left( {{M_1}} \right)$. Finally, ${Y_c}\left( {{M_1}} \right)$ is multiplied with the initial saliency map ${M_0}$. The final saliency map $L_{Abs - CAM}^c$ is obtained after the ReLU operation.

\begin{equation}
	\centering
L_{Abs - CAM}^c = {\mathop{\rm Re}\nolimits} LU\left( {\sum\limits_k {{Y_c}\left( {{M_1}} \right){M_0}}} \right) 
\end{equation}

The details of proposed Abs-CAM are given in the following Algorithm 1.

\begin{algorithm}[htbp]
\caption{The proposed Abs-CAM interpretable algorithm}
\hspace*{0.01in} {\bf Input:} 
Image ${X_0}$, Model ${Y}$, Class ${c}$ \\
\hspace*{0.01in} {\bf Output:} ${L_{Abs - CAM}^c}$
\begin{algorithmic} [1]
\State Get target convolutional layer ${l}$, feature map ${A_l^k}$, score ${Y_c\left(X_0 \right)}$;
\State ${C} {\leftarrow}$ the number of channels in ${A_l^k}$;
\For{${k}$ in [0, ..., ${C-1}$]}
  \State Gradients ${\partial_{ij}^{kc}}$ ${\leftarrow}$ Backpropagation;
    \State  Add the absolute value$:$ $|\partial _{ij}^{kc}|$;
    
    \State  Weight ${W_k^c}$ ${\leftarrow}$ Global-average-pool; 
    \State  The initial saliency map  ${{M_0} = S\left( {U\left(    {A_l^k.W_k^c} \right)} \right)}$; 
    \State  ${M_1} {\leftarrow} {M_0}  \cdot  {X_0}$;
    \State  Softmax get the score: ${{Y_c}\left({{M_1}} \right){M_0}}$;
\EndFor
\State \Return ${L_{Abs - CAM}^c = {\mathop{\rm Re}\nolimits} LU\left({\sum\limits_k {{Y_c}\left({{M_1}} \right){M_0}}} \right)}$
\end{algorithmic}
\end{algorithm}

\section{Experimental results and discussion} \label{4}
In this section, we experimentally evaluate the effectiveness of the proposed method. First, we qualitatively evaluate the proposed method by visualizing saliency maps in Section 4.1. The faithfulness evaluation is following in Section 4.2. We further quantitatively evaluate the methods using the metrics Pointing Game and Sanity Check in Section 4.3 and 4.4, respectively. 

In the following experiments, the pre-trained VGG16 network is used as the base model by default. And the publicly available object classification dataset, namely, ILSVRC2012 val is used in our experiment, which contains 50000 images. For all images in the dataset, they are first resized to the shape (224 * 224 * 3) and transformed to the range [0, 1], and then normalized using mean vector [0.485, 0.456, 0.406] and standard deviation vector [0.229, 0.224, 0.225].

\subsection{Qualitative Evaluation via Visualization}
\subsubsection{Visual Comparison}
\begin{figure}[h]
	\includegraphics[width=3.3in]{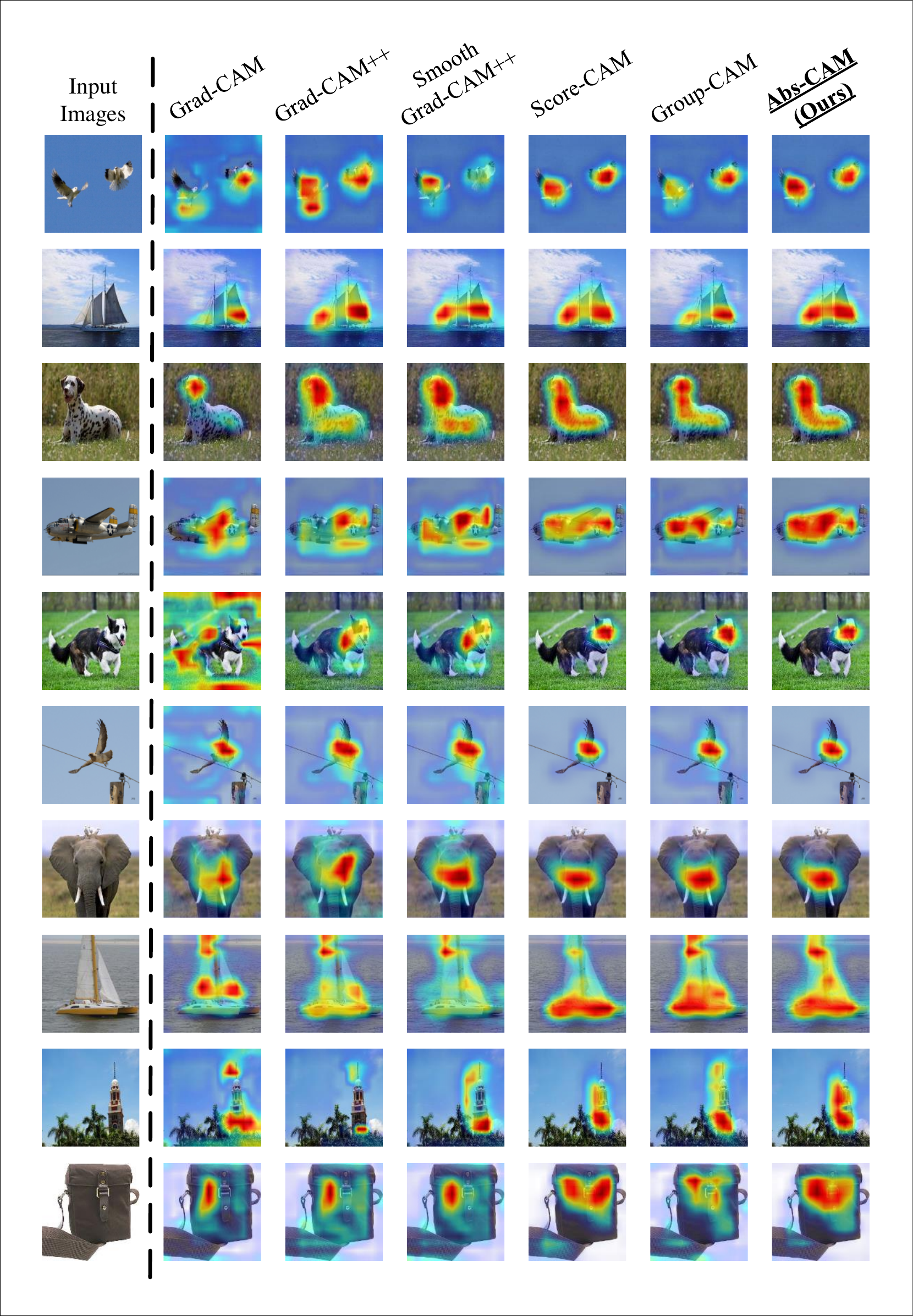}
	\caption{Visual comparison of different class activation mapping-based methods.}
	\label{fig2}
\end{figure}

We qualitatively compares our method with five class activation mapping-based methods by visualizing the saliency maps. Results are shown in Fig. \ref{fig2}, the saliency maps generated by Abs-CAM significantly have less random noises, more concentrated target regions, and better class discrimination compared with three gradient-based CAM methods, including Grad-CAM \cite{selvaraju2017grad}, Grad-CAM++ \cite{chattopadhay2018grad}, Smooth GradCAM++ (SGCAM++) \cite{omeiza2019smooth}. 

The main difference between Abs-CAM and  three gradient-based CAM methods is the weight acquisition method. Grad-CAM, Grad-CAM++ and Smooth GradCAM++ obtain the weight by performing global average pooling on gradients, high-order derivation and adding random noise respectively, while Abs-CAM sets the absolute value of the gradients to enhance the visual features of output neuron activations. The gradient-free Score-CAM\cite{wang2020score} visually outperforms the three gradient-based methods with less noise because the gradient-based method introduces noise when backpropagating to obtain the gradient. However, because Abs-CAM optimizes the gradient, it has a positive impact on the predicted target category, so the obtained saliency area is more concentrated than Score-CAM. Both Abs-CAM and Group-CAM \cite{2021Group} have a stage 2 module, but Abs-CAM optimizes the gradient in the stage 1 module, which makes the visualization effect of Abs-CAM better than Group-CAM, further verifying that the optimized gradient can enhance the visual effect of the saliency map.

\subsubsection{Different CNN Model}
\begin{figure}[h]
	\includegraphics[width=3.3in]{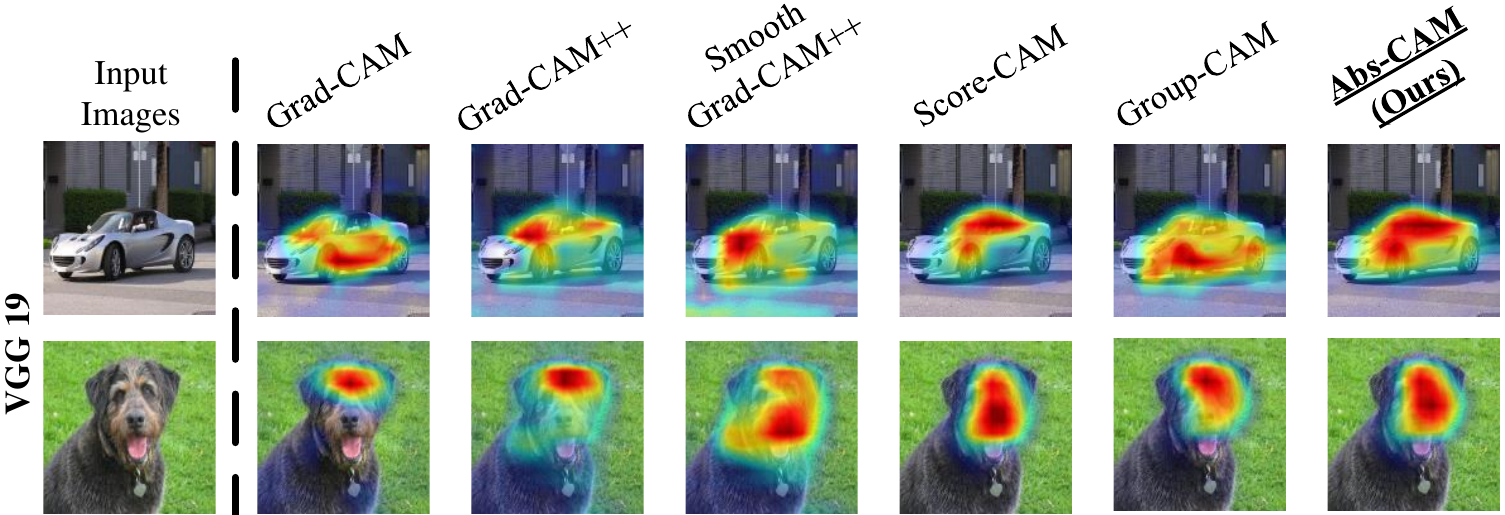}
		\caption{Visual comparison of different CNN.}
	\label{fig3}
\end{figure}

In order to check whether the propose method can be adapted to different CNN structures, a visual comparison is made between the proposed method and other CAM-based methods under VGG19. As shown in Fig. \ref{fig3}, Abs-CAM also shows excellent localisation on VGG19, leading to better visual interpretation results. Therefore, Abs-CAM can be applied to different CNN structures.

\subsubsection{Different classes}
To investigate whether the proposed method in this paper can distinguish different target classes on the same image, we selecte images with two target classes for visual evaluation. As shown in Fig. 4, the VGG16 model classifies the input as ``bull mastiff" with 47.84$\%$ confidence and ``tiger cat" with 0.31$\%$  confidence. Although the prediction probability of the latter is much lower than that of the former, Abs-CAM correctly gives the interpreted positions of the two categories, indicating that Abs-CAM can distinguish between different categories on the same image.

\begin{figure}[h]
	\centering
	\includegraphics[width=2.5in]{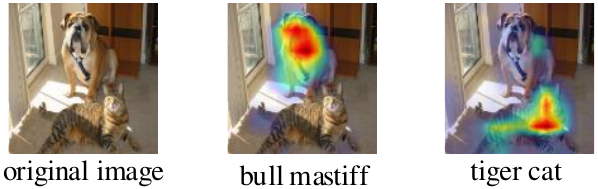}
	\caption{{Verify the ability to distinguish between categories on the same image.}}
	\label{fig4}
\end{figure}

\subsubsection{Comparison of initial and final saliency map}
To verify the improvement of the final saliency map over the initial saliency map, we have visually compared them and notated the initial saliency map as Abs-CAM\_init, the results are shown in Fig. 5. Compared to Abs-CAM\_init, Abs-CAM has a more focused location area with less noise and better visual interpretation. This is due to the addition of the Phase 2 module to Abs-CAM\_init, which further optimises the weighting of the saliency map.

\begin{figure}[htbp]
	\centering
	\includegraphics[width=3.3in]{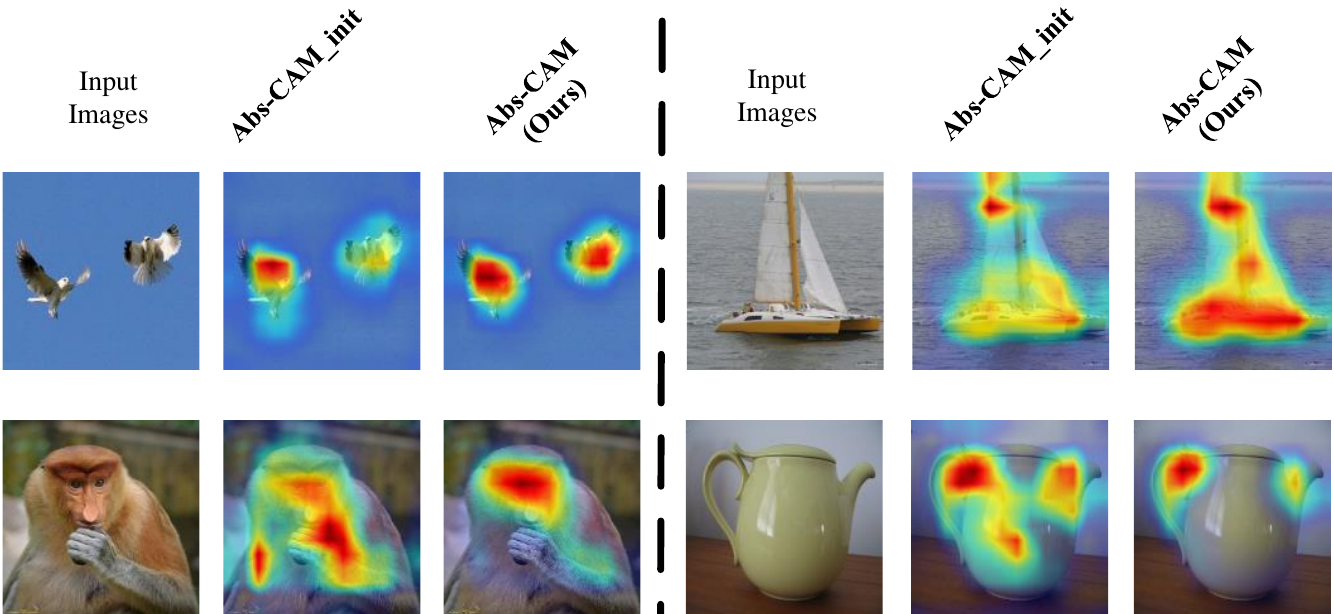}
	\caption{{Visual comparison of final saliency map and initial saliency map.}}
	\label{fig5}
\end{figure}

\subsection{Faithfulness Evaluation}
We first evaluate the objective faithfulness of Abs-CAM through Average Drop and Average Increase as adopted in \cite{chattopadhay2018grad}. The original input is masked by pointwise multiplication with the saliency maps to observe the score change on the target class. In this experiment, top 50$\%$ pixels of the attention map are used as the mask.

We compare the Average Drop and Average Increase results of Abs-CAM, Abs-CAM\_init and five class activation mapping based methods. The above methods use 500 images randomly selected from the ILSVRC2012 val. As shown in Table 2, Abs-CAM achieves 34.2$\%$ average drop and 30.1$\%$ average increase respectively, and outperforms other methods. A good performance reveals that Abs-CAM can successfully find out the most distinguishable region of the target object.

\begin{table}[htbp]
		\centering
		\caption{Average Drop (\%), Average Increase (\%), Insertion and Deletion for different methods.}
		\begin{tabular}{lcccc}
			\hline
			Methods & Drop  & Increase & Insertion & Deletion \bigstrut\\
			\hline
			Grad-CAM & 41.5  & 20.8  & 0.4626 & 0.1110 \bigstrut[t]\\
			Grad-CAM++ & 40.8  & 22.3  & 0.4484 & 0.1179 \\
			SGCAM++ & 41.1  & 23.4  & 0.4504 & 0.1169 \\
			Score-CAM & 35.6  & 29.5  & 0.4929 & 0.1099 \\
			Group-CAM & 35.7  & 29.7  & 0.493 & 0.1108 \\
			Abs-CAM\_init & 39.8  & 25.6  & 0.4712 & 0.1107 \\
			Ours & \textbf{34.2}  & \textbf{30.1}  & \textbf{0.4949} & \textbf{0.1096} \bigstrut[b]\\
			\hline
		\end{tabular}%
		\label{tab2}
\end{table}% 

\begin{figure*}[htbp]
	\centering
	\includegraphics[width=5.5in]{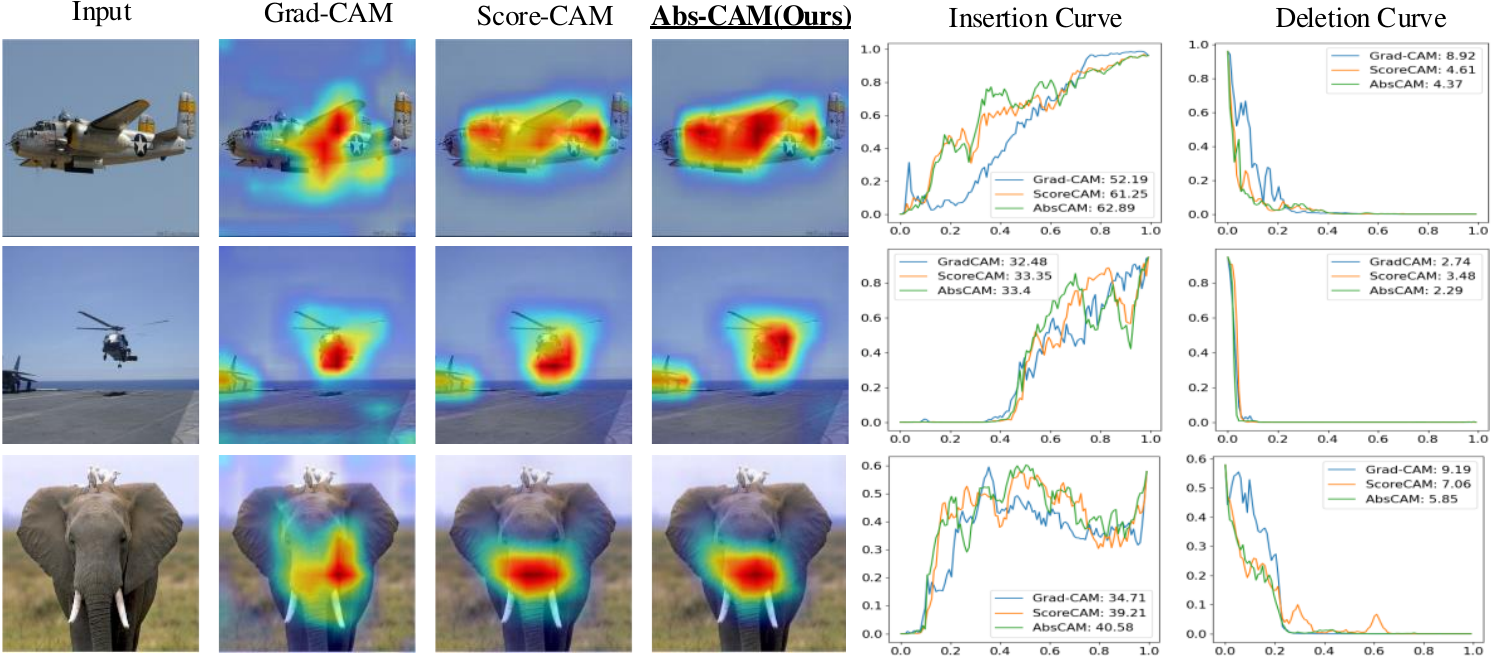}
	\caption{{Comparison of the three methods of deleting and inserting values.}}
	\label{fig6}
\end{figure*}

Furthermore, We conduct the Deletion and Insertion \cite{petsiuk2018rise} metric test, which is used to measure the ability of the interpretable method to capture important pixels. In Deletion, some of the most important pixels are gradually removed and the decreasing value of the prediction probability is calculated until no pixels remain. Therefore, the lower the value of Deletion means that the more important pixels are deleted, and thus the most important pixels can be highlighted accurately. Instead, in Insertion, it starts with a blurred image and gradually introduces pixels until the original image is restored. The more important pixels are introduced leads to a significant increase in the prediction probability of the model, so a higher value of Insertion means that the most important pixels are introduced.

An example of experimental results is shown in Fig. 6, and Table 2 is the average result of 5500 images. As shown in Fig. 6, Abs-CAM outperforms Grad-CAM \cite{selvaraju2017grad} and Score-CAM \cite{wang2020score} in terms of both visual effects and deletion and insertion values. As shown in Table 2, Abs-CAM shows better performance in both metrics and further validates that adding the stage 2 module can improve the performance of the interpretable method. Because Abs-CAM has more concentrated saliency regions, less noise, and larger gradient values of identified significant pixels, Abs-CAM performs the lowest deletion values when significant pixels are removed. Similarly, Abs-CAM shows the highest insertion value when significant pixels are inserted.
 
We record the running times of our method and the five class activation mapping-based methods on the above 5500 images. The results are shown in Table 3. Our method takes a little longer when running 5500 images, but it is in the same order of magnitude compared to other methods. Our method requires more running time due to the use of the phase 2 module, but achieves better performance in terms of visual effects and deletion and insertion values.

\subsection{Pointing Game}
In this section, we use Pointing Game to measures the quality of the saliency map through localization ability. And this metric requires manually labeled bounding boxes as auxiliary information in the calculation process. Specifically, for a saliency map, pointing accuracy is defined as the ratio of the number of points that fall within the bounding box to the number of all points. That is

\begin{equation}
A{\rm{cc}} = \frac{{\# H{\rm{its}}}}{{\# H{\rm{its}} + \# M{\rm{isses}}}}
\end{equation}
where \# H{\rm{its}} is the number of points falling within the bounding box, and \# M{\rm{isses}} is the number of points falling outside the bounding box. 

We use 300 images to compare the pointing accuracy results of our method, Abs-CAM\_init and five methods based on class activation mapping. The results are shown in Table 3. The pointing accuracy value of Abs-CAM is higher than other methods. The saliency regions of Abs-CAM are more concentrated and less noisy, and therefore a higher probability of being hit within the border.

\begin{table}[htbp]
	\centering
	\caption{Pointing accuracy and running times of different methods.}
	\begin{tabular}{lcc}
		\hline
		\textbf{Methods} & \textbf{Accuracy} & \textbf{Time(s)} \bigstrut\\
		\hline
		Grad-CAM & 0.589 & \textbf{16.84} \bigstrut[t]\\
		Grad-CAM++ & 0.576 & 17.83 \\
		SGCAM++ & 0.594 & 22.43 \\
		Score-CAM & 0.637 & 22.65 \\
		Group-CAM & 0.642 & 17.14 \\
		Abs-CAM\_init & 0.601 & 16.98 \\
		Abs-CAM(ours) & \textbf{0.652} & 23.46 \bigstrut[b]\\
		\hline
	\end{tabular}%
	\label{tab3}%
\end{table}%

\subsection{Sanity Check}
Julius \cite{adebayo2018sanity} et al. point out that relying on visual assessment alone may be misleading, and propose an assessment metric--Sanity Check to test whether the interpretable method is sensitive to the model parameters. We use Cascade Randomization (CR) and Independent Randomization (IR) methods are used to compare the output of Abs-CAM on a pre-trained VGG16. As shown in Fig.7, the sanity check results of Abs-CAM with different convolution layers are different. Therefore, Abs-CAM passes the sanity check and its result is sensitive to model parameter and can reflect the quality of model.

\begin{figure}[htbp]
	\centering
	\includegraphics[width=2.5in]{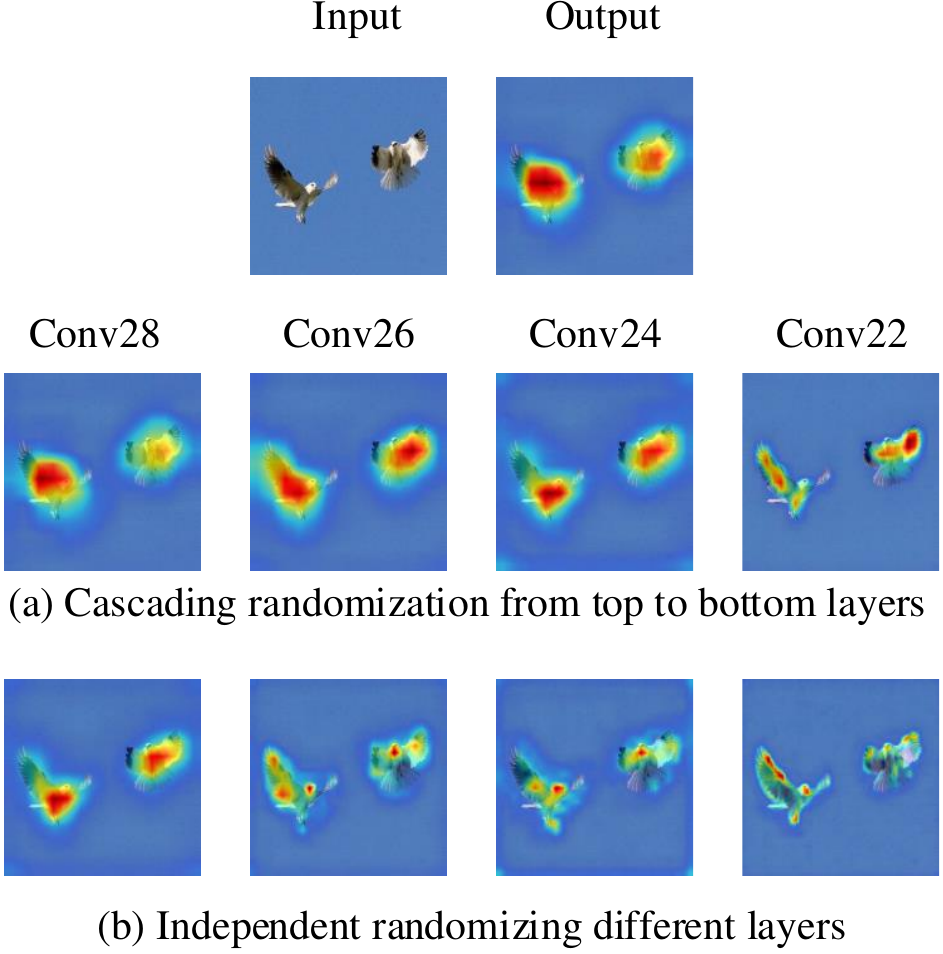}
		\caption{{Sanity check results by cascade randomization and independent randomization.}}
	\label{fig7}
\end{figure}

\section{Conclusion} \label{5}
In this paper, we proposes a noveal gradient optimization interpretable method, Abs-CAM. It adopts the strategy of taking the absolute value to turn all the gradients obtained by backpropagation into positive values, and further optimizes the weight of the initial saliency map. As a result, it effectively reduces the noise of the saliency map, target regions of interest is located more intensively and the saliency map has better class discrimination. Since the initial saliency map is further processed and fed into the model, it increases the complexity of the system and increases the time consumption. In addition, Abs-CAM is also a gradient-based class activation mapping method, which cannot avoid the noise caused by backpropagation. Therefore, future work requires further denoising operations to reduce redundant information at the edges of saliency map regions. 

\section*{Acknowledgements}
This work was supported by National Natural Science Foundation of China (No. 61901165, No. 62177022, and No. 61501199), Collaborative Innovation Center for Informatization and Balanced Development of K-12 Education by MOE and Hubei Province (No. xtzd2021-005), Self-determined Research Funds of CCNU from the Colleges’ Basic Research and Operation of MOE (No. CCNU20ZT010), and Hubei Natural Science Foundation (No. 2017CFB683).

\bibliographystyle{spmpsci}
\bibliography{bmc_article}

\end{document}